\documentclass[runningheads]{llncs}
\usepackage{graphicx}

\usepackage{amsmath,amsfonts}
\usepackage{algorithmic}
\usepackage{algorithm}
\usepackage{array}
\usepackage{textcomp}
\usepackage{stfloats}
\usepackage{url}
\usepackage{verbatim}
\usepackage{graphicx}
\usepackage{amssymb}
\usepackage{booktabs}
\usepackage{bm}
\usepackage{threeparttable}
\usepackage{multirow} 
\usepackage{color}
\usepackage{cite}
\begin{document}
\title{InstancePin: Instance-Addressable Layout-to-Image Diffusion via Coordinate Pinning}
\titlerunning{InstancePin}

\author{
	Chaoyue Wu \inst{1} \and
	Yunfei Zhang\inst{2} \and
	Si Wu\inst{1}\thanks{Corresponding author.}
}

\authorrunning{C. Wu et al.}

\institute{
	School of Computer Science and
	Engineering, South China University of Technology, Guangzhou, Guangdong,
	P. R. China \\
	\email{w1094943231@gmail.com, cswusi@scut.edu.cn}
	\and
	School of Future Technology, South China University of Technology, Guangzhou, Guangdong,
	P. R. China\\
	\email{csyfzhang@gmail.com}
}

\maketitle              
\begin{abstract}

Layout-to-image diffusion models have achieved impressive semantic controllability by conditioning generation on category-level segmentation maps. However, such category-aligned control is not necessarily instance-addressable: multiple nearby objects from the same category are often treated as a shared semantic region, leading to ambiguous boundaries, averaged appearances, and feature confusion among instances. This limitation is particularly evident in urban scene synthesis, where small and crowded pedestrians or vehicles require fine-grained instance separation while preserving global scene consistency. In this paper, we propose InstancePin, an instance-addressable layout-to-image diffusion framework that pins each object instance with an explicit coordinate anchor. Instead of directly injecting instance masks into the pretrained backbone, InstancePin introduces an independent instance-aware adapter to preserve the category-level generation prior while learning instance-specific spatial control. For each instance, its center coordinate is encoded with Fourier features and projected into a coordinate token, which serves as a spatial anchor queried by latent image features through coordinate pinning attention. To make these anchors spatially meaningful, we further supervise the coordinate attention maps with instance regions, encouraging each coordinate token to activate its corresponding object area. Finally, an instance-mask guided fusion module routes pretrained backbone features to non-instance regions and adapter features to instance regions, enabling local instance refinement without sacrificing global semantic fidelity. Extensive experiments on Cityscapes demonstrate that InstancePin mitigates instance entanglement in dense layouts and improves both image fidelity and semantic consistency.

\end{abstract}

\section{Introduction}

Layout-to-image generation aims to synthesize realistic images from spatial layouts such as semantic segmentation maps. Recent diffusion-based methods have substantially improved the fidelity and controllability of this task by combining pretrained text-to-image diffusion models with additional spatial conditions \cite{zhang2026classbooth}. In particular, category-level segmentation maps provide explicit semantic regions for different object classes, allowing the model to generate images that follow the global scene structure. Despite the progress, category-level layout control remains insufficient for scenes containing multiple nearby instances from the same class. A semantic segmentation map indicates where a category appears, but not how same-category instances should be separated during generation. As a result, several adjacent pedestrians or vehicles may be treated as a shared semantic region. The generated images often suffer from ambiguous boundaries, averaged local appearances, and feature confusion among neighboring instances. The problem becomes more pronounced in dense urban scenes, where pedestrians are usually small, crowded, and visually similar. Improving instance-level separation while preserving category-level semantic consistency is therefore a key challenge for layout-to-image diffusion.

\begin{figure*}
	\vspace{-0.5cm}
	\centering
	\includegraphics[width=1.\linewidth]{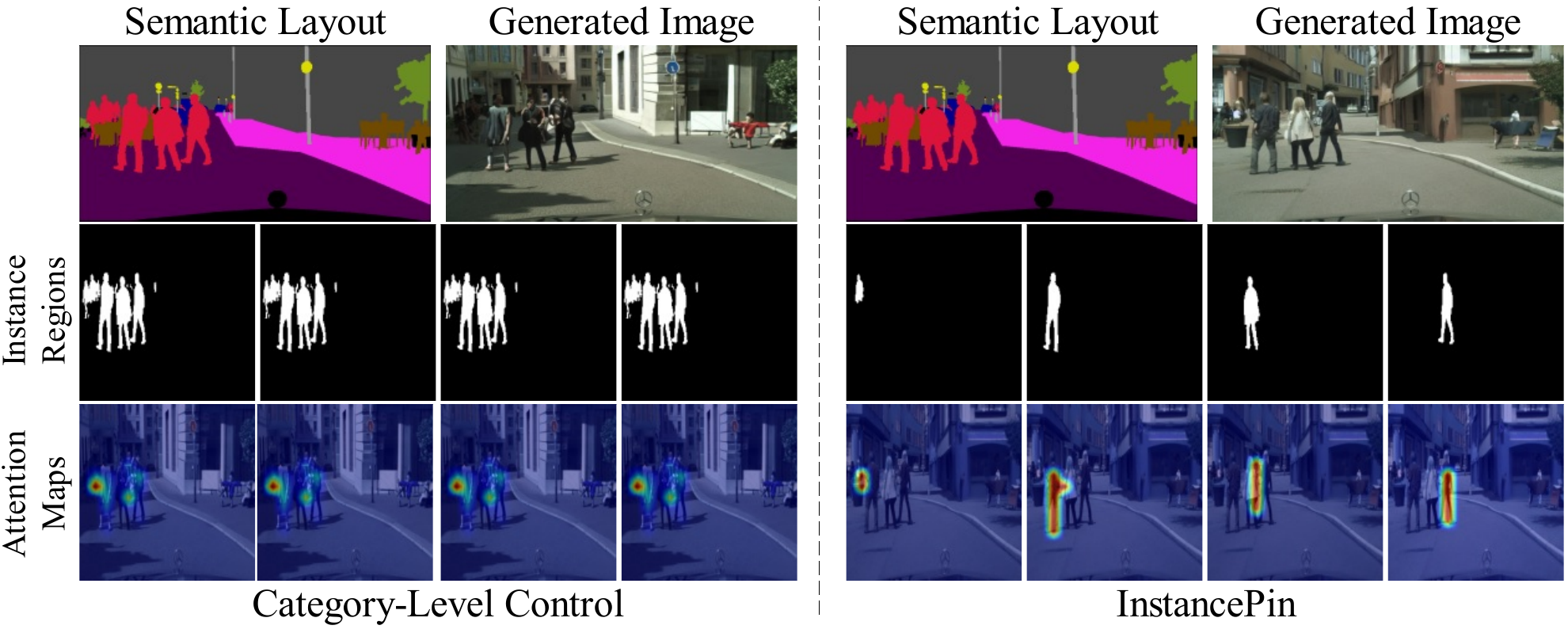}
	\vspace{-0.3cm}
	\caption{Illustration of instance entanglement and disentanglement in layout-to-image diffusion. Given the same semantic layout, category-level control often produces overlapping responses for adjacent pedestrians, while InstancePin utilizes coordinate anchors to activate instance-specific regions and generate distinguishable local structures.}
	\label{fig:motivation}
	\vspace{-0.5cm}
\end{figure*}

Existing controllable diffusion models mainly focus on injecting category-level spatial conditions into pretrained diffusion backbones. ControlNet~\cite{ControlNet} and Uni-ControlNet~\cite{Uni-ControlNet} introduce additional control branches to encode external conditions such as semantic maps, enabling strong layout consistency. FreestyleNet~\cite{FreestyleNet} further rectifies cross-attention maps so that each text token affects its corresponding semantic region. Although these methods improve spatial alignment between layouts and generated images, the control signal is still primarily organized around semantic categories. Multiple instances from the same category share the same semantic token or mask region, making it difficult to assign distinct generation behavior to each instance. As illustrated in Fig.~\ref{fig:motivation}, category-level control tends to produce entangled responses for nearby pedestrians, while InstancePin encourages instance-specific activation through coordinate anchors.

To address this limitation, we propose InstancePin, an instance-addressable layout-to-image diffusion framework that improves instance-level spatial separation through coordinate pinning. The central idea of InstancePin is to preserve the category-level generation ability of the pretrained diffusion backbone while introducing a separate pathway for fine-grained instance control. To this end, we design an Instance-Aware Adapter (IAA) that learns instance-aware adapter features independently from the pretrained backbone. The adapter allows instance-level information to be modeled without directly overwriting the semantic generation prior of the original model. Within the adapter, we further introduce Coordinate Pinning Attention (CPA) to distinguish different instances from the same semantic category. Since mask labels alone are not compact queryable conditions, CPA encodes each instance center with Fourier features and projects it into a coordinate token. Latent image features query these coordinate tokens during denoising, allowing different same-class instances to be associated with distinct spatial anchors. To make the learned anchors spatially meaningful, coordinate attention maps are supervised by the corresponding instance regions, encouraging each coordinate token to activate its own object area rather than nearby instances. Finally, we introduce Instance-Mask Guided Fusion (IMGF) to combine category-level and instance-level features according to their spatial responsibilities. The pretrained backbone features are preserved in non-instance regions to maintain global scene structure, while instance-aware adapter features are injected into instance regions to refine local object details. This fusion module enables InstancePin to improve instance-level generation without sacrificing the semantic consistency provided by the pretrained layout-to-image model. Experiments on Cityscapes show that InstancePin effectively reduces spatial entanglement in dense urban layouts and improves image fidelity and semantic consistency, especially in crowded pedestrian regions. The main contributions of this paper are summarized as follows:

\begin{itemize}
	
	\item Distinct from existing category-level layout control methods, InstancePin targets instance-addressable layout-to-image diffusion by separating nearby same-class objects with coordinate-guided spatial anchors.
	
	\item Coordinate Pinning Attention is introduced to encode instance center coordinates into coordinate tokens and align coordinate attention maps with instance regions, enabling fine-grained association between latent image features and individual instances.
	
	\item We design an instance-mask guided fusion module to integrate pretrained backbone features and instance-aware adapter features, improving local instance generation while preserving global semantic consistency.

\end{itemize}

\section{Related work}
\subsection{Generic image-to-image translation}

Generic image-to-image translation maps images from a source domain to a target domain while preserving semantic consistency. Conditional GANs \cite{ConditionalGAN,GAN,zhang2025class} enabled controllable synthesis by introducing conditional inputs into adversarial learning. Based on this paradigm, Pix2Pix \cite{Pix2Pix} established supervised image translation with encoder-decoder architectures, and Pix2PixHD \cite{Pix2PixHD} improved high-resolution synthesis through coarse-to-fine generation and multi-scale discriminators. Later methods extended image translation to more flexible settings. StarGAN \cite{StarGAN} and DiscoGAN \cite{DiscoGAN} explored multi-domain translation, MUNIT \cite{MUNIT} decoupled content and style for multimodal generation, and CycleGAN \cite{CycleGAN} introduced cycle consistency for unpaired translation. Recently, diffusion-based image translation has shown strong generation quality, with PITI \cite{PITI} demonstrating that pretrained diffusion models can be adapted to various downstream translation tasks.

\subsection{Semantic image synthesis}

Semantic image synthesis generates realistic images from semantic masks and is closely related to layout-to-image generation. SPADE \cite{SPADE} introduced spatially-adaptive normalization to preserve semantic information in normalization layers, while GLADE \cite{GLADE} further improved class-wise control. Other works improved efficiency or spatial adaptivity through group convolutions and kernel prediction \cite{GroupDNet,SCGAN}. In addition to category-level masks, INADE \cite{INADE} introduced instance-level modulation to improve diversity. OASIS \cite{OASIS} improved semantic consistency by using a segmentation-based discriminator. More recently, diffusion models \cite{Diffusion-model} have reshaped semantic image synthesis. DP-SIMS \cite{DP-SIMS} leverages pretrained image backbones, and SDM \cite{SDM} integrates spatially-adaptive normalization into diffusion models. Despite these advances, most semantic synthesis methods still rely primarily on category-level layouts or instance-level modulation, without explicitly associating diffusion features with individual same-category instances.

\subsection{Controllable diffusion models}

Controllable diffusion models introduce additional conditions to guide generation. Early text-to-image generation models such as GLIDE \cite{GLIDE}, DALL-E \cite{DALL-E}, and DALL-E2 \cite{DALL-E2} used language or CLIP \cite{CLIP} embeddings to guide image synthesis. LDM \cite{LDM} reduced computational cost by performing diffusion in the VAE latent space, and Stable Diffusion variants have been widely applied to image generation and editing tasks \cite{DETEX,W-plus-adapter}. Beyond text prompts, spatial conditions have been introduced for finer control. Spatext \cite{Spatext} explored spatio-textual representations, FreestyleNet \cite{FreestyleNet} rectified cross-attention maps to associate text tokens with semantic regions, and PLACE \cite{PLACE} used layout control maps to improve semantic consistency. Efficient conditioning mechanisms have also been studied. GLIGEN \cite{GLIGEN} adds learnable attention layers, T2I-Adapter \cite{T2I-Adapter} extracts spatial features for frozen diffusion backbones, ControlNet \cite{ControlNet} uses a copied branch to encode external controls, and Uni-ControlNet \cite{Uni-ControlNet} unifies local and global controls with feature denormalization.

Overall, existing semantic image synthesis and controllable diffusion methods have made substantial progress in category-level layout alignment. However, same-category instances are often represented by shared semantic masks or shared text tokens, leaving instance-level spatial separation underexplored in diffusion-based layout-to-image generation. InstancePin differs from these methods by assigning each instance a coordinate anchor and learning coordinate attention maps that associate latent image features with individual instance regions. The resulting instance-aware adapter features are further fused with pretrained backbone features through instance-mask guided fusion, allowing local instance refinement while preserving category-level semantic consistency.

\begin{figure}[htp]
	\centering
	\includegraphics[width=1.\linewidth]{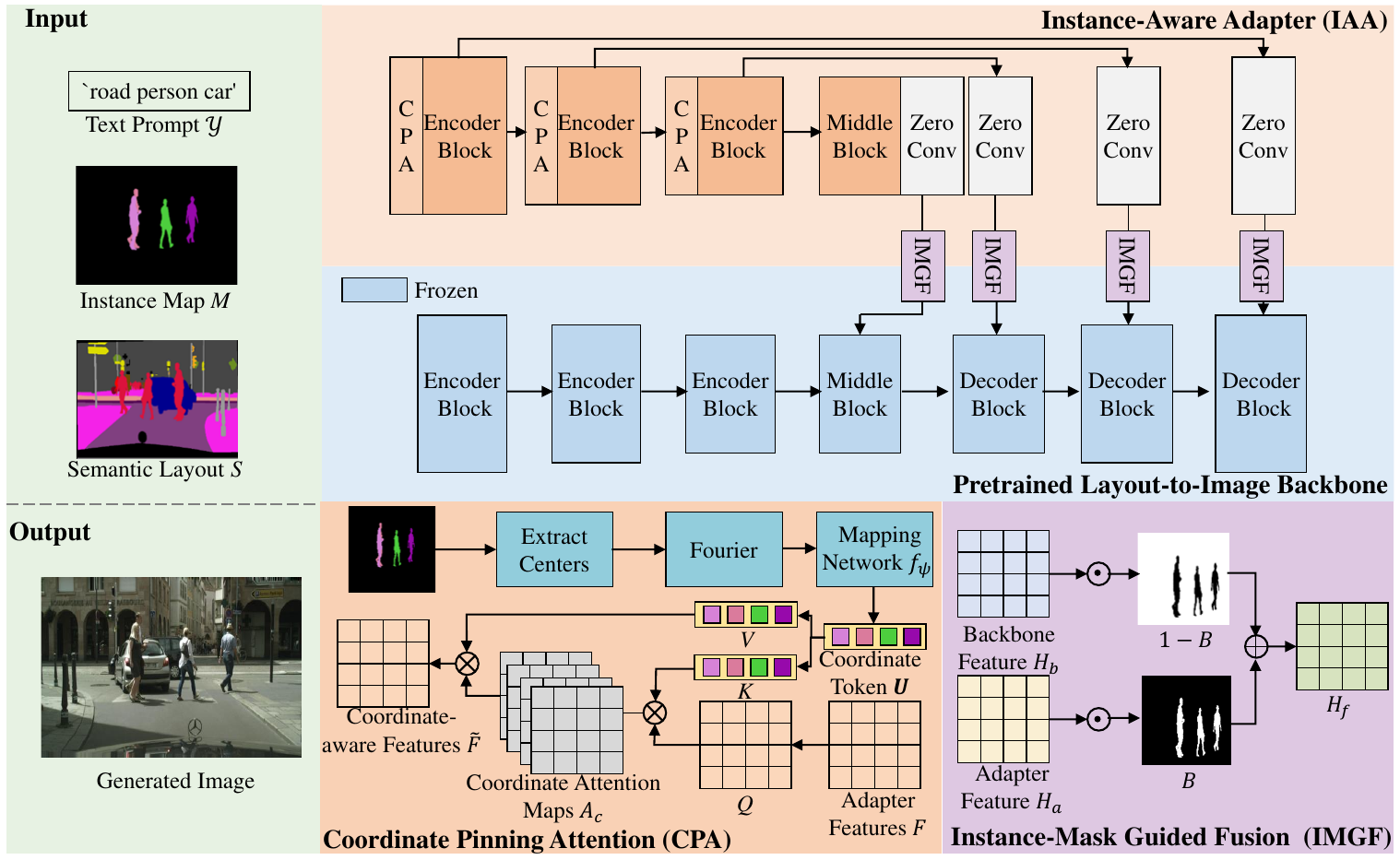}
	\caption{An overview of the proposed InstancePin. To enable instance-level spatial separation, InstancePin extracts instance centers $P$ and a binary instance mask $B$ from the instance map $M$. The instance-aware adapter learns multi-scale adapter features independently from the pretrained layout-to-image backbone. Within coordinate pinning attention, instance centers are encoded into coordinate tokens $U$ to associate adapter features with individual instance regions. Furthermore, the instance-mask guided fusion module routes pretrained backbone features $H_b$ and adapter features $H_a$ according to $B$, producing fused features $H_f$ for instance-aware generation.}
	\label{fig:framework}
\end{figure}

\section{Proposed Method}

\subsection{Overview}

Given a semantic layout $S$ and a text prompt $y$, layout-to-image diffusion models synthesize an image by progressively denoising a latent variable under category-level spatial guidance. The semantic layout specifies where each category should appear, but adjacent instances from the same category are still represented by shared semantic control. InstancePin introduces instance-level spatial anchors to separate nearby same-class objects without changing the category-level generation role of the pretrained denoising backbone. Let $M$ denote the instance map, where pixels belonging to the $k$-th instance have the same instance index. From $M$, we derive two forms of instance information: a coordinate set $P=\{p_k\}_{k=1}^{K}$ and a binary instance region mask $B$. Here $p_k=(x_k,y_k)$ is the center coordinate of the $k$-th instance, $K$ is the number of valid instances, and $B$ marks all instance regions regardless of instance index. The coordinates $P$ are used to construct compact instance anchors, while $B$ is used for region-wise feature fusion. The overall architecture of InstancePin is illustrated in Fig. \ref{fig:framework}.

\subsection{Instance-Aware Adapter}

The pretrained denoising backbone has learned strong category-level layout priors, and directly fine-tuning all its layers with instance annotations may weaken such priors. InstancePin therefore introduces an Instance-Aware Adapter as an independent branch for instance-level spatial control. The adapter mirrors the encoder and middle-block structure of the denoising U-Net and can be initialized from the corresponding pretrained weights. Its multi-scale outputs are projected by zero-initialized convolutional layers (ZeroConv), which stabilize fine-tuning by keeping the initial adapter contribution close to zero.

For a noisy latent $z_t$ at timestep $t$, the Instance-Aware Adapter takes the text prompt $y$, semantic layout $S$, and coordinate set $P$ as conditions, and produces multi-scale adapter features:
\begin{equation}
	\{H_a^l\}_{l=1}^{L}
	=
	\mathcal{A}_{\phi}(z_t,t,y,S,P),
\end{equation}
where $\mathcal{A}_{\phi}$ denotes the adapter, $\phi$ represents its learnable parameters, and $l$ indexes the feature scale. The adapter features are designed to provide instance-aware refinements rather than replace the pretrained denoising process globally. 

\subsection{Coordinate Pinning Attention}

Instance regions provide accurate object extents, but raw instance indices are not suitable as queryable conditions for diffusion features. Coordinate Pinning Attention represents each instance with a coordinate token derived from its center coordinate. The token acts as a spatial anchor that can be queried by latent image features during denoising.

For each center coordinate $p_k$, Fourier features are computed as follows:
\begin{equation}
	\gamma(p_k)=
	[
	\sin(\omega_1 p_k), \cos(\omega_1 p_k), \dots,
	\sin(\omega_F p_k), \cos(\omega_F p_k)
	],
\end{equation}
where $\{\omega_f\}_{f=1}^{F}$ are predefined frequency bands. A learnable coordinate mapping network then projects the Fourier feature into a coordinate token: $u_k=f_{\psi}(\gamma(p_k))$, where $u_k \in \mathbb{R}^{d_c}$ and $f_{\psi}$ denotes the mapping network. The coordinate tokens of all valid instances are collected as: $U=[u_1,u_2,\dots,u_K]$.

Given an adapter feature $F^l \in \mathbb{R}^{N_l \times d_l}$ at scale $l$, where $N_l=H_lW_l$, latent positions query the coordinate tokens through cross-attention. The coordinate attention map is computed as follows:
\begin{equation}
	A_c^l
	=
	\operatorname{Softmax}
	\left(
	\frac{(F^l W_Q^l)(U W_K^l)^{\top}}{\sqrt{d}}
	\right),
\end{equation}
where $W_Q^l$ and $W_K^l$ are learnable projections, $d$ is the attention dimension, and $A_c^l \in \mathbb{R}^{N_l \times K}$ describes the association between latent spatial locations and instance coordinate tokens. The coordinate-aware feature is obtained as: $\tilde{F}^l=A_c^l(UW_V^l)$, where $W_V^l$ is the value projection. Invalid padded coordinates are masked before the softmax operation, so attention is computed only over valid instances.

To prevent coordinate tokens from becoming weak or ambiguous conditions, InstancePin explicitly aligns coordinate attention maps with instance regions. The instance map is resized to the attention resolution and converted into an instance-region target $R^l \in \{0,1\}^{N_l \times K}$, where $R^l_{i,k}=1$ indicates that latent position $i$ belongs to the $k$-th instance. The coordinate alignment loss is computed as follows:

\begin{equation}
	\mathcal{L}_{\mathrm{coord}}
	=
	\mathbb{E}_{z_0,t,M}
	\left[
	\frac{1}{|\Omega|}
	\sum_{l \in \Omega}
	\left\|
	A_c^l(z_t,t,P) - R^l
	\right\|_2^2
	\right],
\end{equation}
where $\Omega$ denotes the set of scales supervised by instance regions. The target $R^l$ is obtained from the instance map paired with $z_0$ and resized to the resolution of $A_c^l$. This supervision encourages each coordinate token to activate its corresponding object area and suppress responses from neighboring instances.

\subsection{Instance-Mask Guided Fusion Module}

After Coordinate Pinning Attention produces instance-aware adapter features, Instance-Mask Guided Fusion Module combines them with features from the pretrained denoising backbone according to spatial responsibility. The pretrained features are more reliable for global category-level structure and non-instance regions, whereas the adapter features are used to refine regions where instance separation is required.

At scale $l$, let $H_b^l$ be the feature from the pretrained denoising backbone and $H_a^l$ be the corresponding adapter feature. The binary instance mask $B$ is resized to the same spatial resolution and denoted as $B^l$. Feature fusion is performed as follows:
\begin{equation}
	H_f^l
	=
	(1-B^l) \odot H_b^l
	+
	B^l \odot H_a^l,
\end{equation}
where $H_f^l$ is the fused feature and $\odot$ denotes element-wise multiplication with channel-wise broadcasting. Non-instance regions therefore keep the pretrained backbone features, while instance regions receive adapter features. This region-wise fusion avoids globally perturbing the pretrained denoising process and concentrates instance-aware generation capacity on crowded object regions.

\subsection{Model Optimization}

InstancePin is optimized with diffusion denoising supervision and coordinate attention alignment. Given a clean latent $z_0$, Gaussian noise $\epsilon$, and timestep $t$, the forward diffusion process produces $z_t=\sqrt{\bar{\alpha}_t}z_0+\sqrt{1-\bar{\alpha}_t}\epsilon$. The complete denoising model, denoted as $\epsilon_{\theta}$, predicts the noise under text, semantic layout, coordinate anchors, and instance-region fusion:
\begin{equation}
	\mathcal{L}_{\mathrm{diff}}
	=
	\mathbb{E}_{z_0,\epsilon,t}
	\left[
	\left\|
	\epsilon -
	\epsilon_{\theta}(z_t,t,y,S,P,B)
	\right\|_2^2
	\right],
\end{equation}
where $\theta$ includes all trainable parameters of InstancePin. The optimization objective combines the denoising loss with coordinate alignment:
\begin{equation}
	\mathcal{L}
	=
	\mathcal{L}_{\mathrm{diff}}
	+
	\lambda_{\mathrm{coord}}
	\mathcal{L}_{\mathrm{coord}},
\end{equation}
where $\lambda_{\mathrm{coord}}$ balances the two terms. Through this objective, InstancePin learns coordinate-aware instance control while retaining the semantic layout generation ability of the pretrained diffusion model.

\section{Experiments}

\subsection{Experimental Settings}

\textbf{Datasets and Implementation Details.} We conduct experiments on the Cityscapes dataset, which contains 3,475 urban street scene images with a resolution of 2048$\times$1024. Following the standard split, 2,975 images are used for training and 500 images are used for testing. All images and layouts are resized to 512$\times$512 for training and inference. Cityscapes provides both category-level semantic layouts and instance-level annotations. In addition to the full validation set, we report results on a pedestrian-focused subset, denoted as Cityscapes-pedestrian, to examine generation quality in crowded same-category instance regions where spatial entanglement is more likely to occur. InstancePin is built upon the pretrained FreestyleNet. The Instance-Aware Adapter and coordinate mapping network are optimized with AdamW. The learning rate is set to $1\times10^{-5}$, the batch size is set to 2, and the weight $\lambda_{\mathrm{coord}}$ in Eq. (7) is set to 10. All experiments are conducted on an NVIDIA RTX 3090 GPU.

\textbf{Evaluation Metrics.} We evaluate generated images from two aspects. For image fidelity, we use FID (Frechet Inception Distance) \cite{heusel2017gans}, which measures the distribution distance between generated images and real images in the Inception feature space. For semantic consistency, we use a pretrained DRN-D-105 \cite{yu2017dilated} segmentation model to parse generated images and compute mIoU (Mean Intersection over Union) between the predicted segmentation maps and the ground-truth semantic layouts. Lower FID indicates better image fidelity, while higher mIoU indicates stronger semantic consistency.

\subsection{Comparison with Existing Methods}

We compare InstancePin with representative semantic image synthesis and controllable diffusion methods, including OASIS, SDM, ControlNet, Uni-ControlNet, and FreestyleNet. The comparison is conducted on both the full Cityscapes validation set and the pedestrian-focused subset. The full validation set evaluates overall layout-to-image generation quality, while the pedestrian-focused subset emphasizes dense same-category regions that are central to our problem setting.

As shown in Table~\ref{tab:sota}, InstancePin achieves the best performance on both FID and mIoU across the two evaluation settings. On the full Cityscapes validation set, InstancePin improves FID from 40.3 to 37.6 and mIoU from 73.3 to 75.1 compared with FreestyleNet, the strongest baseline in the table. On Cityscapes-pedestrian, InstancePin further improves FID from 46.6 to 43.5 and mIoU from 74.1 to 75.8 over FreestyleNet. The consistent gains on both the full set and the pedestrian-focused subset indicate that instance-level control improves crowded object regions while maintaining global image fidelity and semantic consistency.

\begin{figure}[h]
	\centering
	\includegraphics[width=1.\linewidth]{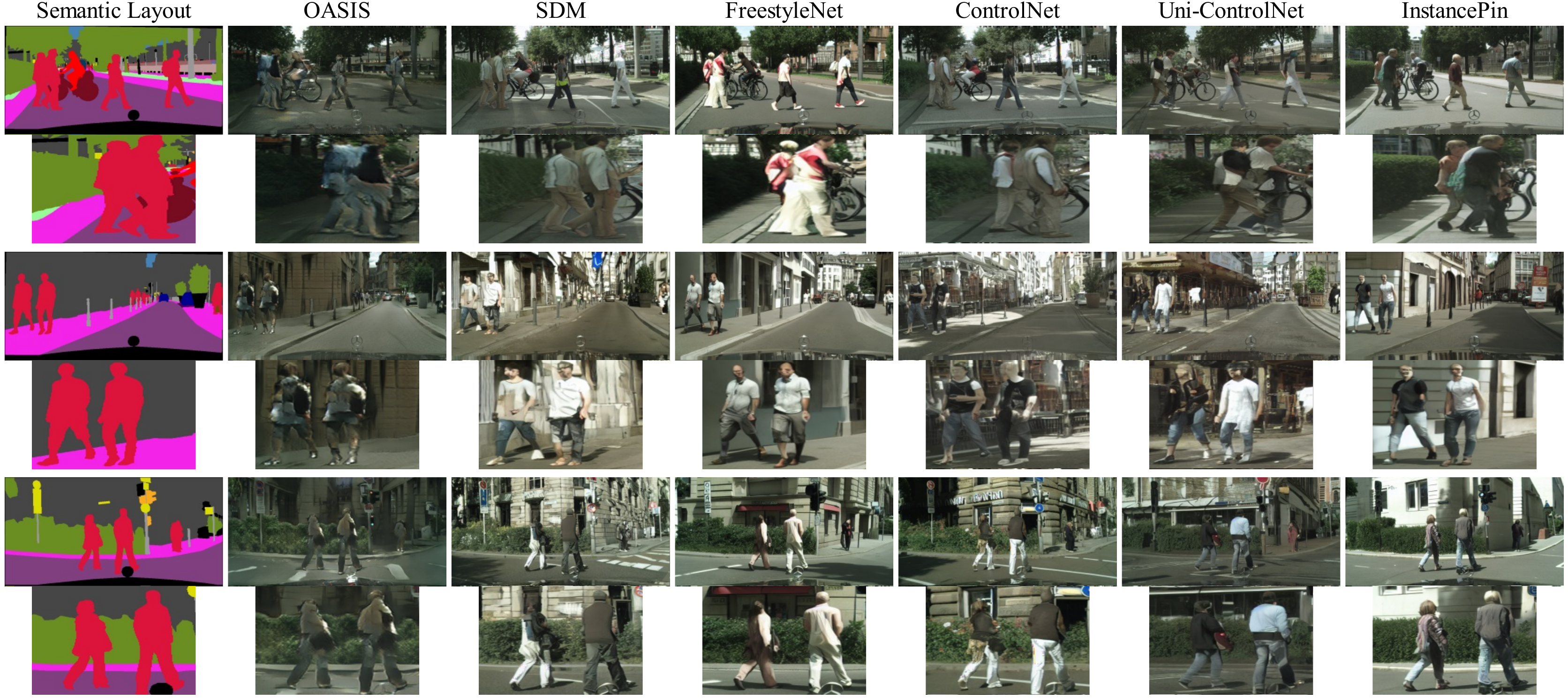}
	\caption{Qualitative comparison with competing methods on Cityscapes. Each group shows the generated image and the zoomed pedestrian region.}
	\label{fig:compare}
\end{figure}

Furthermore, Fig.~\ref{fig:compare} presents qualitative comparisons under the same semantic layouts. Existing methods are generally able to follow the coarse category-level layout, but adjacent pedestrians may still be rendered with blurred boundaries, merged local structures, or similar appearances. InstancePin produces clearer separations between nearby pedestrians and preserves more distinguishable local details in the zoomed regions. This visual trend is consistent with the quantitative improvements on the pedestrian-focused subset.

\begin{table}[ht]
	\centering
	\caption{Comparison of InstancePin with competing models in terms of image quality and semantic consistency.}
	\resizebox{.6\linewidth}{!}{
		\begin{tabular}{l c c c c c}
			\toprule
			
			& \multicolumn{2}{c}{Cityscapes} & & \multicolumn{2}{c}{Cityscapes-pedestrian} \\ 
			
			\cmidrule(lr){2-3} \cmidrule(lr){5-6}

			Method & FID $\downarrow$ & mIoU $\uparrow$ & & FID $\downarrow$ & mIoU $\uparrow$ \\ 
			\midrule
			OASIS          & 47.7 & 68.5 & & 52.8 & 68.7 \\
			SDM            & 42.1 & 69.2 & & 49.5 & 69.6 \\
			FreestyleNet   & 40.3 & 73.3 & & 46.6 & 74.1 \\
			ControlNet     & 41.1 & 72.1 & & 47.3 & 72.7 \\
			Uni-ControlNet & 44.3 & 71.5 & & 51.2 & 72.3 \\
			\midrule 
			\textbf{InstancePin}  & \textbf{37.6} & \textbf{75.1} & & \textbf{43.5} & \textbf{75.8} \\
			\bottomrule
		\end{tabular}
	}
	\label{tab:sota}
\end{table}

\begin{figure}[h]
	\centering
	\includegraphics[width=0.95\linewidth]{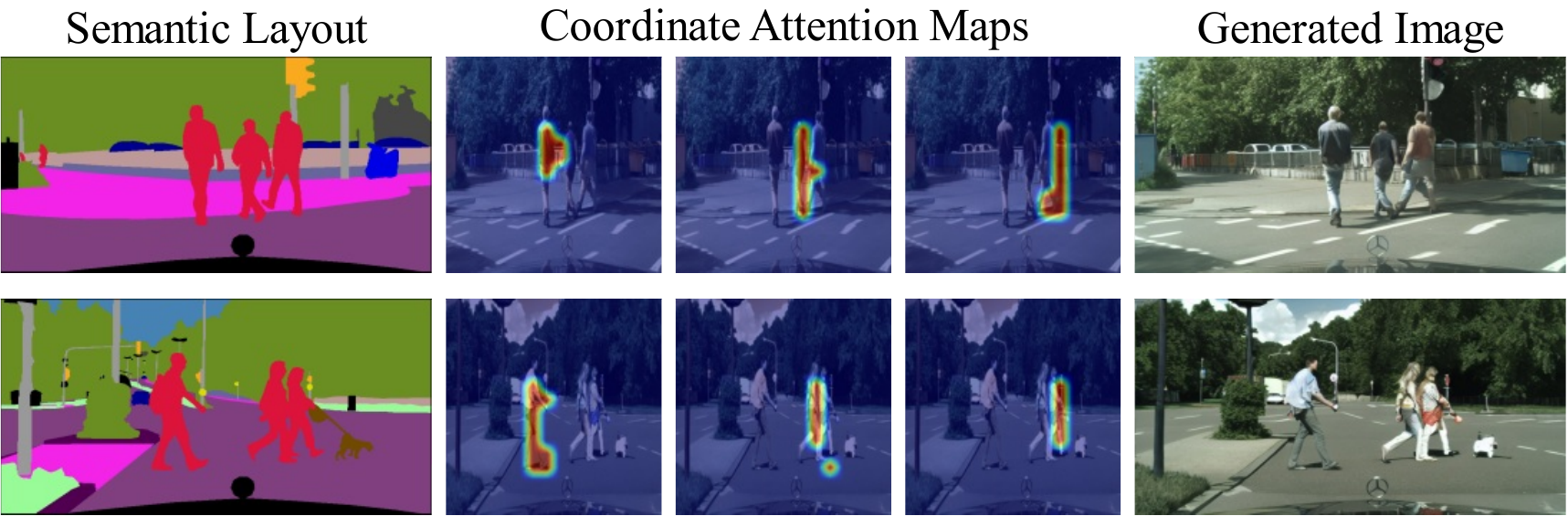}
	\caption{Visualization of coordinate pinning attention. Each row shows the coordinate attention maps associated with different coordinate tokens and the corresponding generated result.}
	\label{fig:cca}
\end{figure}

\subsection{Analysis of Coordinate Pinning Attention}

To examine whether coordinate tokens learn meaningful spatial correspondence with individual instances, we visualize the coordinate attention maps during inference. Each coordinate token is generated from the center coordinate of one instance, and its attention map reflects the spatial locations that query this token during denoising. As shown in Fig.~\ref{fig:cca}, the attention response of each coordinate token is concentrated around the corresponding instance region. Responses to neighboring instances or background regions are suppressed, even when multiple instances belong to the same semantic category and appear close to one another. This observation suggests that coordinate pinning attention provides a spatially interpretable mechanism for associating latent image features with individual instances, which helps reduce same-category instance entanglement.

\begin{table}[h]
	\centering
	\caption{Quantitative comparison of InstancePin and different variants.}
	\resizebox{.6\linewidth}{!}{
		\begin{tabular}{l c c c c c}
			\toprule
			& \multicolumn{2}{c}{Cityscapes} & & \multicolumn{2}{c}{Cityscapes-pedestrian} \\ 
			\cmidrule(lr){2-3} \cmidrule(lr){5-6}
			
			Method & FID $\downarrow$ & mIoU $\uparrow$ & & FID $\downarrow$ & mIoU $\uparrow$ \\ 
			\midrule
			w/o $\mathcal{L}_{\mathrm{coord}}$    & 43.2 & 71.9 & & 49.6 & 72.3 \\
			w/o IMGF  & 39.3 & 73.8 & & 45.1 & 74.3 \\
			\midrule 
			InstancePin    & \textbf{37.6} & \textbf{75.1} & & \textbf{43.5} & \textbf{75.8} \\
			\bottomrule
		\end{tabular}
	}
	\label{tab:ablation}
\end{table}

\subsection{Ablation Study}

\begin{figure}[h]
	\centering
	\includegraphics[width=1.\linewidth]{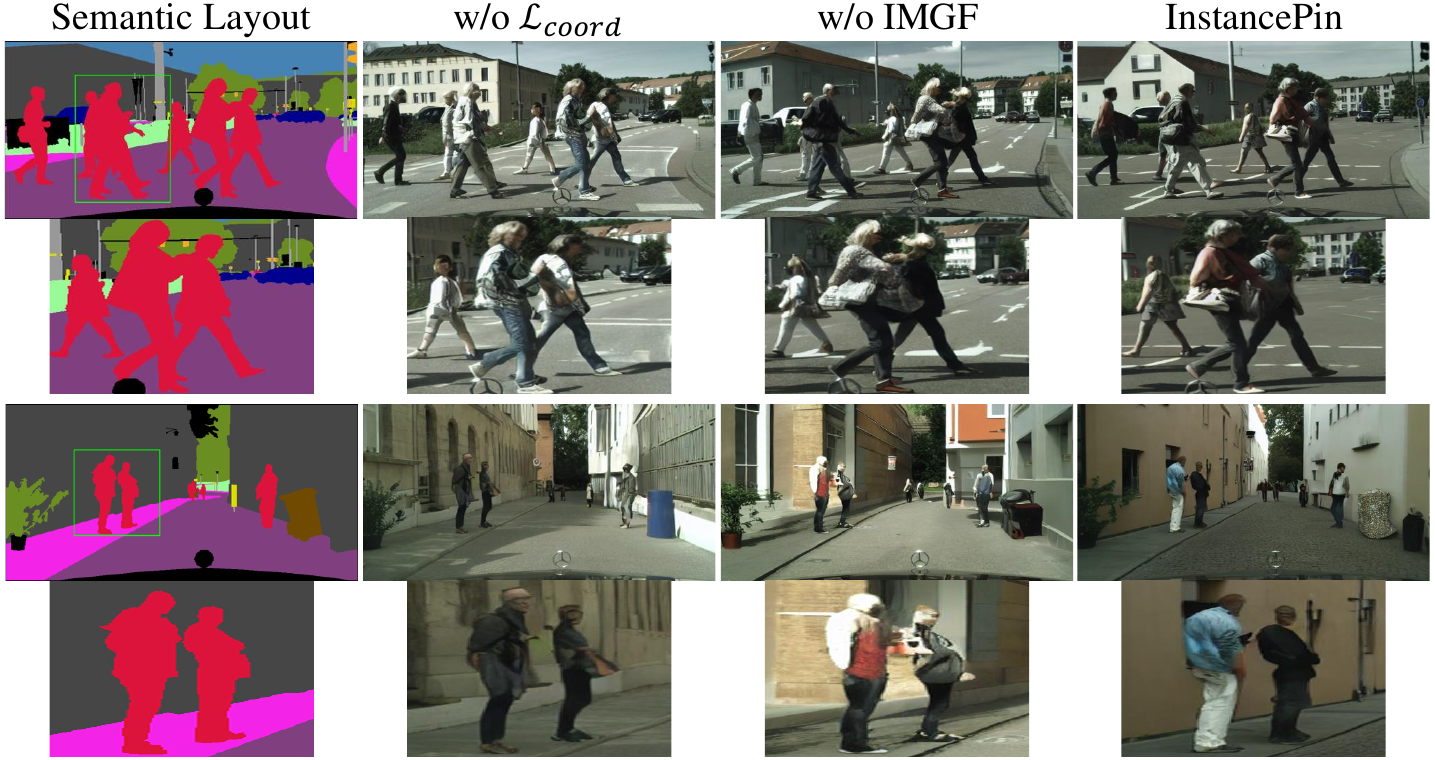}
	\caption{Qualitative ablation study of the main components in InstancePin.}
	\label{fig:ablation}
\end{figure}

We conduct ablation studies to evaluate the contribution of the main components in InstancePin. The variants are compared on both the full Cityscapes validation set and the pedestrian-focused subset, as shown in Table~\ref{tab:ablation}. Qualitative results are provided in Fig.~\ref{fig:ablation}.

Removing the coordinate alignment loss weakens the supervision between coordinate attention maps and instance regions. The variant `w/o $\mathcal{L}_{\mathrm{coord}}$' shows clear degradation in both image fidelity and semantic consistency, indicating that coordinate tokens need explicit spatial alignment to serve as reliable instance anchors. The degradation on the pedestrian-focused subset further confirms the importance of coordinate alignment for crowded same-category instances.

The variant `w/o IMGF' removes the instance-mask guided fusion module and combines pretrained backbone features with adapter features without instance-region routing. This variant increases FID from 37.6 to 39.3 and decreases mIoU from 75.1 to 73.8 on the full validation set. On Cityscapes-pedestrian, FID increases from 43.5 to 45.1 and mIoU decreases from 75.8 to 74.3. The performance drop shows that explicit instance-mask guided fusion helps preserve non-instance regions with pretrained backbone features while concentrating adapter features on instance regions.

\section{Conclusion}

We presented InstancePin for instance-addressable layout-to-image diffusion in dense urban scenes. By introducing coordinate anchors for individual instances, InstancePin enables the denoising process to associate latent image regions with nearby same-category objects more explicitly. The instance-aware adapter provides local instance refinements while preserving the category-level generation ability of the pretrained diffusion model, and the instance-mask guided fusion module keeps this refinement concentrated on instance regions. As a result, InstancePin reduces spatial entanglement among adjacent objects and produces clearer local structures without sacrificing global semantic consistency. Experiments on Cityscapes show consistent improvements over existing semantic-layout controlled methods, especially on the pedestrian-focused subset where crowded same-category instances are common. We believe InstancePin provides a practical step toward fine-grained and instance-aware controllable image synthesis under complex spatial layouts.

\bibliographystyle{splncs04}
\bibliography{sample-base}
\end{document}